\documentclass[preprint,review,10pt,times]{elsarticle}
\biboptions{sort&compress}

\usepackage[
  top=4.3cm,
  right=4.8cm,
  bottom=4.3cm,
  left=4.8cm
]{geometry}

\usepackage{amssymb}
\usepackage{amsmath}
\usepackage{amsthm}
\usepackage{bm}
\usepackage{caption}
\usepackage{subcaption}

\usepackage{booktabs}

\usepackage{multirow}
\usepackage{placeins}

\usepackage[hidelinks]{hyperref}

\usepackage{threeparttable}

\newcommand{\equatoref}[1]{\hyperref[#1]{Eq.~(\ref*{#1})}}

\DeclareCaptionLabelFormat{myformat}{\textbf{Fig. #2}}
\DeclareCaptionLabelFormat{mytableformat}{\textbf{Table #2}}
\journal{Pattern Recognition}

\begin{document}

\begin{frontmatter}



\title{\fontsize{14}{16}\selectfont CHOW-SLAM: Compact Hybrid Representation with Complementary Overlap Window Optimization for RGB-D SLAM}


\author[1]{Wenxuan Ji}

\author[1]{Jin Xiao\corref{cor1}}

\author[1]{Xiaoguang Hu}

\author[1]{Jiaqi Shi}

\author[1]{Zichong Jia}

\author[1]{Baochang Zhang}

\affiliation[1]{
    organization={School of Automation Science and Electrical Engineering, Beihang University},
    city={Beijing},
    postcode={100191},
    country={China}
}

\cortext[cor1]{%
\parbox[t]{0.95\linewidth}{%
Corresponding author.\\
\textit{E-mail addresses:}
\href{mailto:jiwenxuan@buaa.edu.cn}{jiwenxuan@buaa.edu.cn} (W. Ji),
\href{mailto:xiaojin@buaa.edu.cn}{xiaojin@buaa.edu.cn} (J. Xiao),
\href{mailto:xiaoguang@buaa.edu.cn}{xiaoguang@buaa.edu.cn} (X. Hu),
\href{mailto:jiazichong@buaa.edu.cn}{jiazichong@buaa.edu.cn} (Z. Jia),
\href{mailto:shijiaqi@buaa.edu.cn}{shijiaqi@buaa.edu.cn} (J. Shi),
\href{mailto:bczhang@buaa.edu.cn}{bczhang@buaa.edu.cn} (B. Zhang).
}}

\begin{abstract}
Simultaneous localization and mapping (SLAM) based on Neural Radiance Fields (NeRF) enables dense, continuous scene reconstruction. However, existing systems operating with limited online resources struggle to simultaneously construct two types of constraints, namely, compact yet discriminative spatial constraints derived from scene representations and persistent temporal constraints derived from historical observations. To address this challenge, we propose CHOW-SLAM, a dense RGB-D SLAM framework that explicitly constructs these complementary spatial and temporal constraints. Spatially, we propose a compact parametric--hash (P-H) hybrid representation that organizes components based on planes and grids across scales in P and H branches. A unified multi-output decoder further aligns the ray termination distributions induced by TSDF and density, preserving geometry and appearance under a compact parameter budget. 
Temporally, we propose a complementary overlap-window strategy to prevent optimization from being dominated by short-term overlap or weakly related historical observations. Within a fixed budget, the strategy retains recent frames, selects high-overlap local frames, and introduces temporally distributed historical keyframes. Loss-aware keyframe insertion and bundle adjustment scheduling further adapt optimization to tracking quality. In addition, ORB-based tracking and geometric pose estimation are used for pose initialization, followed by neural rendering optimization to improve tracking stability. Extensive evaluations on multiple datasets demonstrate that CHOW-SLAM outperforms state-of-the-art methods in both scene reconstruction quality and camera tracking accuracy. The source code is available at \url{https://github.com/jinjidexiaohuoban/CHOW-SLAM}.
\end{abstract}




\end{frontmatter}



\section{Introduction}
\label{sec:introduction}
Visual simultaneous localization and mapping (SLAM) is a fundamental problem in computer vision and robotics, enabling robotic navigation and providing spatial perception for embodied intelligent applications such as vision-language navigation~\cite{wang2025navigation,liu2025embodiednavigation}.
Over the past decades, traditional visual SLAM frameworks achieve high localization accuracy through hand-crafted geometric methods~\cite{orbslam3,vinsmono,lsdslam,lin2025slam2,munoz2020ucoslam}, while learning-based approaches further improve pose-estimation robustness by incorporating learned geometric and semantic priors~\cite{teed2021droidslam,czarnowski2020deepfactors,fan2022blitzslam,xia2026dmsaaslam}. However, sparse or semi-dense maps are insufficient for embodied agents, which need not only to localize themselves but also to perceive scene geometry and reason about spatial relationships. The emergence of Neural Radiance Fields (NeRF) provides a continuous and differentiable representation for dense scene modeling~\cite{mildenhall2020nerf}, and recent radiance-field studies~\cite{lai2025fast} further show its potential for efficient feed-forward reconstruction from sparse observations. 

Despite these advances, a fundamental challenge remains in NeRF-centric RGB-D SLAM: under a limited online optimization budget, the system must construct sufficiently compact, discriminative, and persistent constraints for continual pose--map optimization. Such constraints are determined by two coupled factors: how the scene is represented in 3D space, and which historical observations are selected to reinforce the neural map over time. The former, corresponding to the representation level, determines whether each sampled ray can provide reliable geometric and appearance cues, whereas the latter, corresponding to the observation level, determines whether these cues can be consistently accumulated across local and historical views. Although recent methods have made progress in both representation design and keyframe selection, it remains challenging to simultaneously achieve compact modeling, fine-grained reconstruction, and long-term consistency in continual online optimization.

At the representation level, recent NeRF-based SLAM methods have explored various scene representations~\cite{zhu2022niceslam,johari2023eslam,muller2022instant,sandstrom2023pointslam}. Among them, hash encodings and feature tri-planes are two representative forms of spatial constraints due to their efficient feature querying and dense mapping capability. Hash encodings store multi-resolution features in compact tables and efficiently capture high-frequency details, as adopted by Co-SLAM~\cite{wang2023coslam}, EC-SLAM~\cite{li2026ecslam}, and QQ-SLAM~\cite{jiang2025qqslam}; however, hash collisions may introduce ambiguous constraints and reconstruction artifacts. Feature tri-planes instead decompose 3D features onto orthogonal planes, providing structured spatial support and improving surface completeness in systems such as ESLAM~\cite{johari2023eslam} and PLGSLAM~\cite{deng2024plgslam}, but their fine-grained 3D discriminability is limited under a compact parameter budget.  Although HS-SLAM~\cite{gong2025hsslam} combines hash grids and tri-planes to exploit their complementary advantages, stacking complete encodings increases parameter redundancy. Therefore, the key issue is not simply to combine encodings, but to organize hash and plane features as scale-aware and complementary components, to construct compact, discriminative, and artifact-resistant spatial constraints.

At the observation level, a compact representation can only benefit SLAM if online optimization is driven by effective and persistent historical constraints. Early neural SLAM systems prioritize informative observations using mapping loss, as in iMAP~\cite{sucar2021imap}, or information-guided sampling, as in iSDF~\cite{ortiz2022isdf}. Although these strategies improve local optimization efficiency, they may overemphasize difficult views and provide limited long-term supervision. Later NeRF-centric SLAM methods incorporate historical observations through overlap-based or global keyframe sampling. Overlap-based selection strengthens local geometric constraints but may reduce temporal diversity, as in NICE-SLAM~\cite{zhu2022niceslam}, whereas global sampling covers a broader keyframe set but may introduce weakly related constraints, as in Co-SLAM~\cite{wang2023coslam}. Therefore, the key challenge is not to include more historical frames, but to construct complementary temporal constraints that jointly preserve recent pose continuity, local geometric consistency, and long-term map stability.

Accordingly, we present CHOW-SLAM, a NeRF-based dense RGB-D SLAM framework that constructs compact spatial constraints through scene representation and persistent temporal constraints through online optimization. (1) For compact spatial constraint construction, we propose a parametric--hash (P-H) hybrid representation, which organizes the parametric P branch and the hybrid hash H branch into scale-aware and functionally complementary components. The P branch provides structured spatial support through explicit coarse grids and fine planes, while the H branch captures multi-scale details through coarse hash planes and a fine hash grid. These two branches are integrated through a unified multi-output decoder, which performs self-supervision by aligning the ray termination distributions derived from the truncated signed distance field (TSDF) and density field, thereby preserving detailed geometry and appearance under a compact parameter budget.

(2) For persistent temporal constraint selection, we further propose a complementary overlap-window optimization strategy. Instead of relying on a single source of historical observations, the optimization window considers recent frames, high-overlap local frames, and historical keyframes. These observations respectively preserve short-term pose continuity, provide strong local geometric constraints, and maintain long-term supervision against map forgetting and accumulated drift. Together with loss-aware keyframe insertion and bundle adjustment (BA) scheduling, this strategy strengthens neural map constraints and improves online pose--map optimization.

\begin{figure*}[t]
\centering
\includegraphics[width=0.8\textwidth]{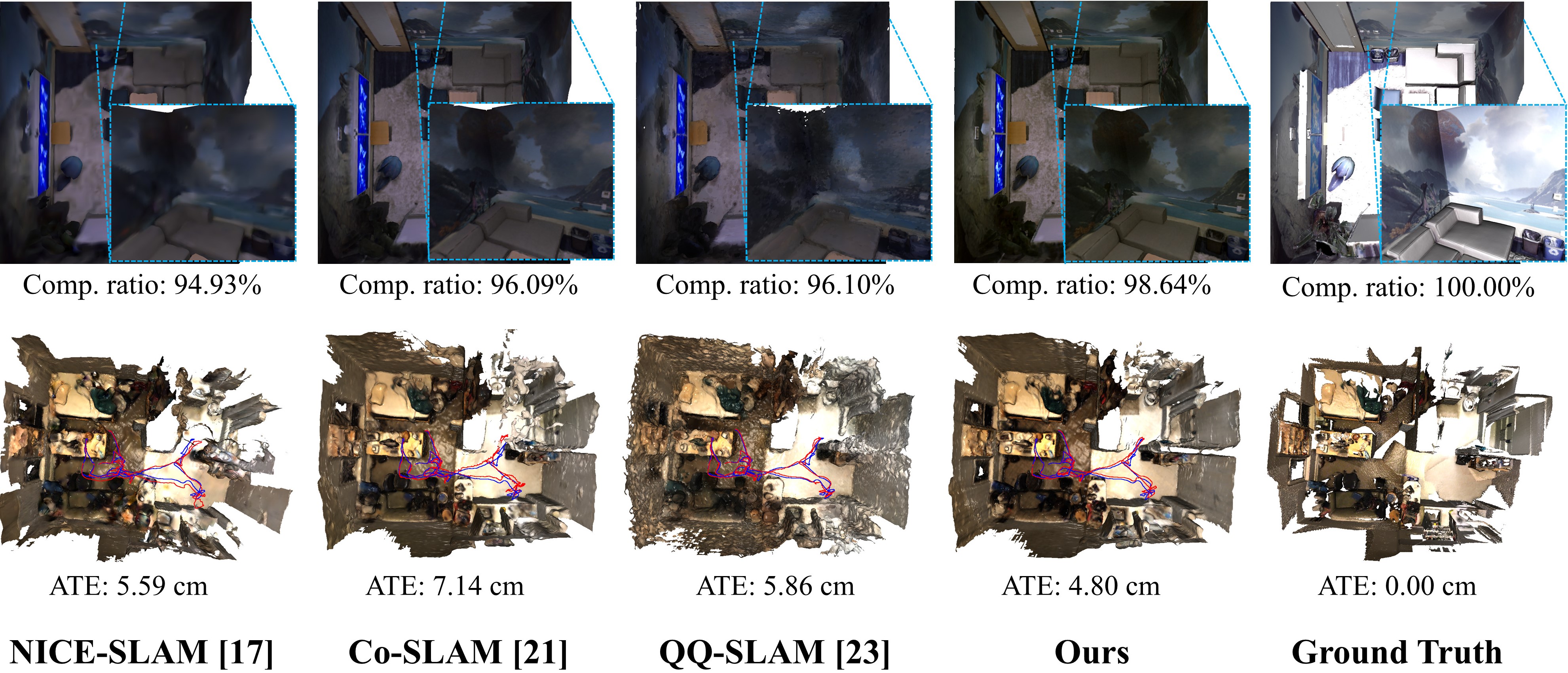}
\caption{Qualitative comparison on Replica~\cite{straub2019replica} office0 and ScanNet~\cite{dai2017scannet} scene0207. Blue and red curves denote ground-truth and estimated trajectories. Our method achieves more complete reconstruction and more accurate tracking than NICE-SLAM~\cite{zhu2022niceslam}, Co-SLAM~\cite{wang2023coslam}, and QQ-SLAM~\cite{jiang2025qqslam}. The enlarged views further show that our method produces more faithful color rendering.}
\label{fig:qualitative_comparison}
\end{figure*}

In addition, (3) NeRF-SLAM commonly formulates camera tracking as image alignment and photometric bundle adjustment, which is sensitive to initialization because of its narrow convergence basin and lack of explicit correspondences~\cite{cartillier2024slaim}. To improve pose initialization, we exploit ORB-based tracking and geometric pose estimation from classical visual SLAM~\cite{murartal2017orbslam2}, and further refine the initialized pose through neural rendering optimization. 

Fig.~\ref{fig:qualitative_comparison} presents qualitative comparisons of reconstruction quality and trajectory tracking. In addition to achieving more complete geometry, CHOW-SLAM produces more faithful color rendering, as highlighted by the zoomed-in regions. Extensive evaluations on multiple benchmark datasets further demonstrate its effectiveness in both dense reconstruction and camera tracking.

The main contributions of this paper are summarized as follows:
\begin{itemize}
\item \textbf{CHOW-SLAM framework.} We propose CHOW-SLAM, a reliable NeRF-based dense RGB-D SLAM system that jointly exploits compact neural mapping, overlap-aware optimization, and classical pose initialization. Extensive experiments demonstrate its effectiveness and robustness in dense reconstruction and camera tracking.

\item \textbf{Compact P-H Hybrid Representation.} We introduce a compact parametric--hash (P-H) hybrid representation module that organizes the parametric P branch and the hybrid hash H branch into scale-aware and complementary components. In addition, a unified multi-output decoder aligns the TSDF-derived and density-derived ray termination distributions, enabling detailed scene reconstruction under a compact parameter budget.

\item \textbf{Complementary Overlap Window Optimization.} We propose a complementary overlap-window optimization strategy with loss-aware keyframe insertion and BA scheduling, which strengthens neural map constraints and improves the stability of online pose-map optimization.

\item \textbf{Two-stage Neural Tracking.} We introduce a two-stage tracking scheme that uses ORB-based tracking and geometric pose estimation for lightweight pose initialization, followed by neural rendering refinement to improve stability.

\end{itemize}

\section{Related Work}
\label{sec:related_work}
\subsection{Traditional VSLAM Frameworks}
\label{subsec:traditional_vslam}
Traditional visual SLAM methods mainly rely on feature correspondences and geometric optimization for camera tracking and map construction. Early sparse systems such as MonoSLAM~\cite{davison2007monoslam} demonstrated real-time monocular SLAM, and ORB-SLAM2~\cite{murartal2017orbslam2} and ORB-SLAM3~\cite{orbslam3} further improved robustness through feature-based tracking, local mapping, loop closure, and bundle adjustment. However, sparse landmark maps are insufficient for dense scene understanding and geometric reconstruction. To obtain richer geometry, DTAM~\cite{newcombe2011dtam} introduced direct photometric optimization for semi-dense reconstruction, while Kintinuous~\cite{whelan2015kintinuous} extended RGB-D fusion to large-scale dense mapping. Subsequent systems such as ElasticFusion~\cite{whelan2015elasticfusion} and BAD-SLAM~\cite{schops2019badslam} further improved dense mapping robustness and global consistency. Recent learning-enhanced VSLAM methods incorporate semantic segmentation, dynamic-object filtering, and multimodal scene reasoning into traditional SLAM pipelines, as represented by Blitz-SLAM~\cite{fan2022blitzslam}, SLAM2~\cite{lin2025slam2}, and DMSAA-SLAM~\cite{xia2026dmsaaslam}. Overall, traditional VSLAM provides efficient feature matching and geometric pose estimation, but remains limited in continuous dense representation and appearance-aware mapping. Therefore, our system uses ORB-based tracking for lightweight pose initialization and refines the pose through neural rendering optimization.

\subsection{NeRF-centric VSLAM Frameworks}
\label{subsec:nerf_vslam}
Although NeRF-centric SLAM usually has lower tracking accuracy than traditional VSLAM, it provides continuous and differentiable dense scene representations with appearance modeling capability. iMAP~\cite{sucar2021imap} first achieved online joint optimization of an MLP scene representation and camera poses, but its single-MLP architecture limits scalability. NICE-SLAM~\cite{zhu2022niceslam} addresses this limitation with hierarchical feature grids, Vox-Fusion~\cite{yang2022voxfusion} improves scalability with voxel-based implicit fusion, and Point-SLAM~\cite{sandstrom2023pointslam} adopts neural points for flexible local mapping. ESLAM~\cite{johari2023eslam} and Co-SLAM~\cite{wang2023coslam} further promote efficient scene representation by exploring feature tri-planes and sparse hash encodings, respectively. For hash-based representations, EC-SLAM~\cite{li2026ecslam} introduces constrained TSDF hash encoding and joint optimization, while QQ-SLAM~\cite{jiang2025qqslam} improves compactness through query quantization. For tri-plane representations, PLGSLAM~\cite{deng2024plgslam} adopts local tri-planes and local-to-global bundle adjustment (BA) for large-scale indoor scenes, while LRSLAM~\cite{park2024lrslam} uses low-rank tensor decomposition to alleviate the memory growth of plane-based representations. HS-SLAM~\cite{gong2025hsslam} combines hash grids and tri-planes to exploit their complementary advantages. Despite these advances, existing NeRF-based VSLAM methods still face challenges in constructing compact, discriminative, and persistent constraints under limited online optimization budgets.

Different from the above systems, our system constructs complementary constraints from representation and observation perspectives. At the representation level, we introduce a compact P-H hybrid representation that combines coarse hash planes, a fine hash grid, coarse explicit grids, and fine feature planes through scale-aware complementary branches. At the observation level, we adopt a complementary overlap-window optimization strategy with loss-aware keyframe insertion and BA scheduling, which balances recent frames, high-overlap local frames, and historical keyframes for robust online pose--map optimization. These designs highlight the capability of our system in compact dense representation and robust online optimization.

\subsection{3DGS-centric Visual SLAM Frameworks}
\label{subsec:3dgs_vslam}

3D Gaussian Splatting (3DGS)~\cite{kerbl20233dgs} has recently become an efficient explicit representation for photorealistic rendering, motivating its use in VSLAM for dense mapping and view synthesis. Photo-SLAM~\cite{huang2024photoslam} and RTG-SLAM~\cite{peng2024rtgslam} exploit classical tracking frameworks such as ORB-SLAM to provide reliable pose estimation, while optimizing Gaussian maps for photorealistic reconstruction. In contrast, SplaTAM~\cite{keetha2024splatam} and MonoGS~\cite{matsuki2024gaussianslam} directly optimize 3D Gaussian primitives for tracking and mapping from RGB-D or monocular inputs. Recent systems further improve 3DGS-based SLAM, where GS-SLAM~\cite{yan2024gsslam} introduces adaptive Gaussian expansion and coarse-to-fine tracking, and RGD-SLAM~\cite{wang2026rgdslam} improves robustness in dynamic environments. However, 3DGS-centric VSLAM systems still require careful Gaussian management during online optimization, which may affect surface accuracy, increase runtime and memory costs, and hinder deployment in real-time robotic applications.

\section{Method}
\label{sec:method}
\begin{figure}[t]
    \centering
    \includegraphics[width=0.9\linewidth]{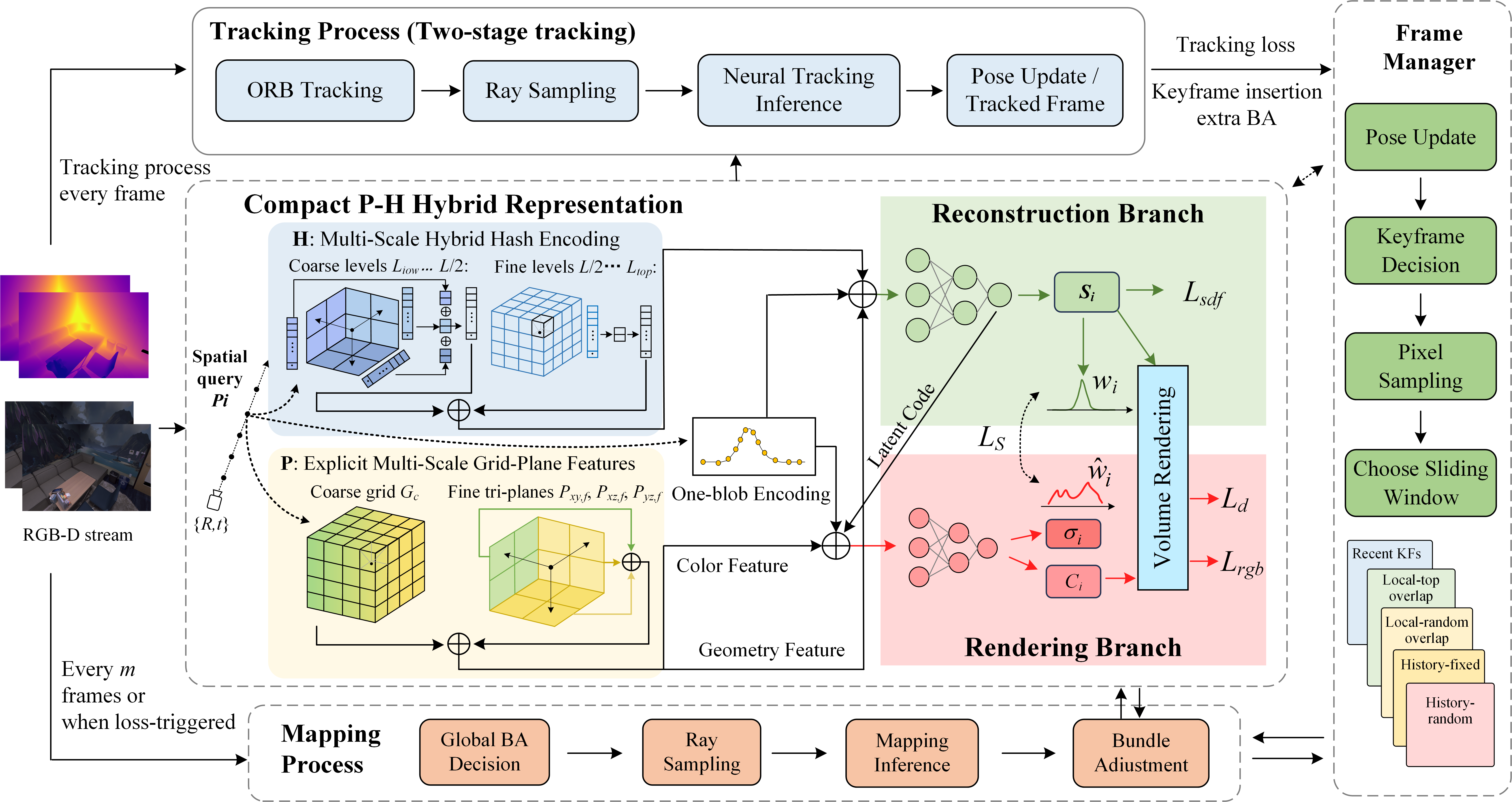}
    \caption{
    Overview of CHOW-SLAM.
    (1) Tracking process: CHOW-SLAM first performs ORB-based tracking to obtain an efficient pose initialization, followed by iterative neural tracking optimization.
    (2) Mapping process: the mapping thread is activated every $m$ frames or when the loss-triggered condition is satisfied, and jointly optimizes the neural scene representation and poses through bundle adjustment.
    (3) Scene representation: the whole scene is modeled by a Compact P-H Hybrid Representation, which consists of H branch and P branch. Two lightweight MLPs in the reconstruction and rendering branches map spatial features to the TSDF $s_i$, volume density $\sigma_i$, and color $\mathbf{c}_i$. $L_c$, $L_d$, $L_{sdf}$, and $L_s$ denote the color loss, depth loss, TSDF loss, and self-supervised consistency loss, respectively.
    (4) Frame manager: the frame manager uses the tracking loss to determine keyframe insertion and construct the complementary sliding window.
    }
    \label{fig:overview}
\end{figure}

Given a sequence of RGB-D frames ${\{I_i,D_i}\}_{i=1}^{M}$ with known camera intrinsics $K\in\mathbb{R}^{3\times3}$, our goal is to estimate the camera poses ${\{\bm{R}_i|t_i\}}_{i=1}^{M}$ and a composite implicit scene representation $F_{\tau}$. The representation jointly models color $\bm{c}$, truncated signed distance $s$, and volume density $\sigma$, where the predicted TSDF field can be further used to extract a 3D mesh through marching cubes. As shown in Fig.~\ref{fig:overview}, CHOW-SLAM consists of two parallel processes, including a tracking process and a mapping process, which share the same neural scene representation. At system initialization, the global map is initialized by performing multiple mapping iterations on the first frame.

For scene representation, we introduce a Compact P-H Hybrid Representation (Sec.~\ref{subsec:compact_ph_hybrid_representation}) to construct compact yet detailed spatial constraints, and design a Multi-output Representation Architecture (Sec.~\ref{subsec:multi_output_representation_architecture}) to jointly predict geometry, density, and color with self-supervised distribution alignment. Sec.~\ref{subsec:volume_rendering} describes the volume rendering process, which converts the predicted scene attributes into rendered color, depth, and SDF observations. For mapping, we propose a Complementary Overlap Window Optimization strategy (Sec.~\ref{subsec:complementary_overlap_window_optimization}) to select recent, high-overlap local, and historical keyframes for persistent temporal constraints. For tracking, we design a Two-stage Neural Tracking scheme (Sec.~\ref{subsec:two_stage_neural_tracking}) to improve pose initialization and tracking stability. Finally, Sec.~\ref{subsec:objective_functions} defines the overall objective functions for joint map and pose optimization.

\subsection{Compact P-H Hybrid Representation}
\label{subsec:compact_ph_hybrid_representation}

Although HS-SLAM~\cite{gong2025hsslam} combines hash grids and tri-planes to exploit their complementary advantages, directly stacking multiple encodings inevitably increases the number of trainable parameters. MHED-SLAM~\cite{feng2026mhedslam} shows that coarse hash planes and a fine hash grid can be combined to balance compactness and detail preservation. This insight naturally raises a further question: whether parametric plane-based representations can also be organized in a scale-aware manner to reduce redundancy while preserving structured spatial constraints. To this end, we introduce a compact parametric--hash (P-H) hybrid representation, where the P branch is derived from the scale-aware design insight in~\cite{feng2026mhedslam} and instantiated as a coarse-grid and fine-plane parametric structure, while the H branch adopts the coarse-plane and fine-grid hash architecture in~\cite{feng2026mhedslam}. In this hybrid design, the P branch provides structured and complementary spatial support to improve representation completeness, the H branch captures fine geometric details with compact multi-scale encoding, and the one-blob encoding serves as a coordinate embedding for low-frequency information~\cite{wang2023coslam}. Together, these components construct compact yet complementary spatial constraints for detailed neural scene representation.

\begin{figure}[t]
    \centering
    \includegraphics[width=0.8\textwidth]{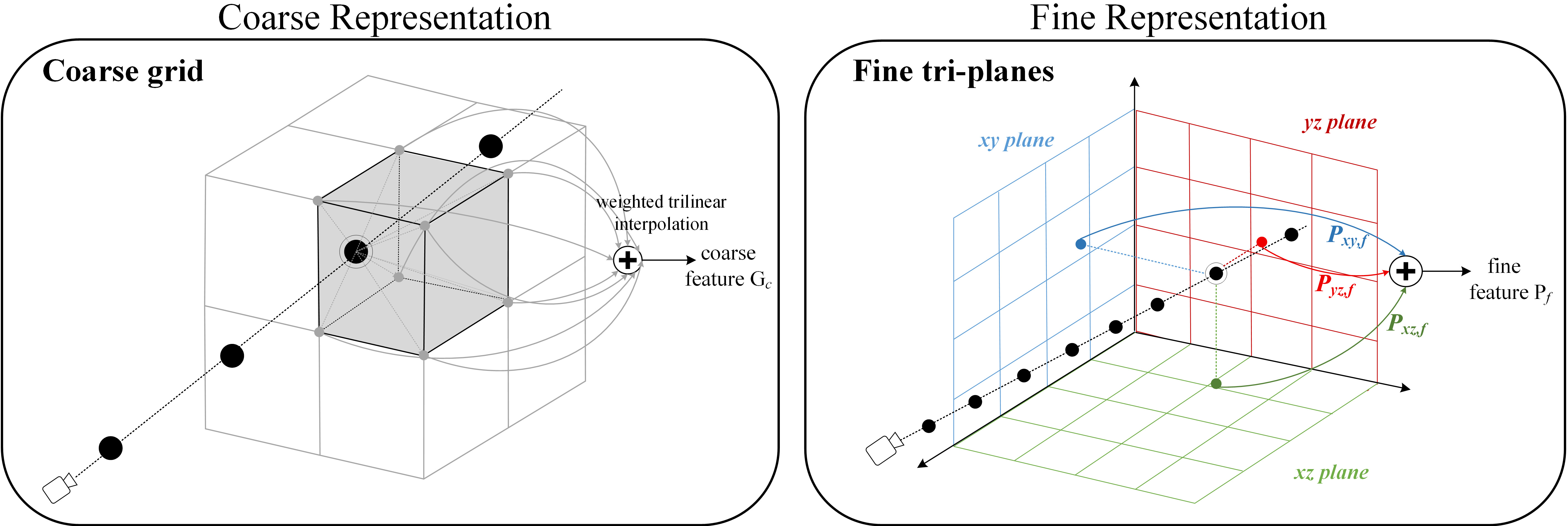}
    \caption{
    Illustration of the parametric P branch. The coarse representation uses an explicit 3D grid queried by weighted trilinear interpolation to provide structured spatial support, while the fine representation uses three orthogonal feature planes queried by bilinear interpolation to efficiently obtain fine-level features.
    }
    \label{fig:p_branch}
\end{figure}

For the parametric P branch, as illustrated in Fig.~\ref{fig:p_branch}, we adopt a coarse-grid and fine-plane design to provide explicit spatial support for geometry and appearance. At the coarse level, the spatial resolution is low, so explicit 3D grids can preserve global structural information with a manageable parameter budget. At the fine level, directly using high-resolution 3D grids would cause cubic parameter growth, whereas three orthogonal feature planes enable high-resolution feature querying with much lower memory cost. Therefore, we use coarse explicit 3D grids to maintain structured spatial constraints and fine feature planes to efficiently capture local details. For a sampled 3D point $\mathbf{x}_i=(x_i,y_i,z_i)$, the geometry feature of the P branch is formulated as
\begin{equation}
\begin{aligned}
\mathbf{P}_c^g(\mathbf{x}_i) &= \mathbf{G}_c^g(\mathbf{x}_i), \\
\mathbf{P}_f^g(\mathbf{x}_i) &= \mathbf{P}_{xy,f}^g(\mathbf{x}_i)
+ \mathbf{P}_{xz,f}^g(\mathbf{x}_i)
+ \mathbf{P}_{yz,f}^g(\mathbf{x}_i), \\
\mathbf{P}^g(\mathbf{x}_i) &= \operatorname{Concat}
\bigl(\mathbf{P}_c^g(\mathbf{x}_i), \mathbf{P}_f^g(\mathbf{x}_i)\bigr).
\end{aligned}
\label{eq:p_branch_geometry_feature}
\end{equation}
Similarly, the appearance feature is defined as
\begin{equation}
\begin{aligned}
\mathbf{P}_c^a(\mathbf{x}_i) &= \mathbf{G}_c^a(\mathbf{x}_i), \\
\mathbf{P}_f^a(\mathbf{x}_i) &= \mathbf{P}_{xy,f}^a(\mathbf{x}_i)
+ \mathbf{P}_{xz,f}^a(\mathbf{x}_i)
+ \mathbf{P}_{yz,f}^a(\mathbf{x}_i), \\
\mathbf{P}^a(\mathbf{x}_i) &= \operatorname{Concat}
\bigl(\mathbf{P}_c^a(\mathbf{x}_i), \mathbf{P}_f^a(\mathbf{x}_i)\bigr).
\end{aligned}
\label{eq:p_branch_appearance_feature}
\end{equation}
where $\mathbf{G}_c^g$ and $\mathbf{G}_c^a$ denote the coarse explicit 3D grids for geometry and appearance, respectively, which are queried by trilinear interpolation. $\mathbf{P}_{xy,f}^{g}$, $\mathbf{P}_{xz,f}^{g}$, and $\mathbf{P}_{yz,f}^{g}$ denote the fine geometry feature planes on the $xy$, $xz$, and $yz$ planes, while $\mathbf{P}_{xy,f}^{a}$, $\mathbf{P}_{xz,f}^{a}$, and $\mathbf{P}_{yz,f}^{a}$ denote the corresponding fine appearance feature planes. These fine feature planes are queried by bilinear interpolation. The coarse and fine features are concatenated to form the final geometry feature $\mathbf{P}^g(\mathbf{x}_i)$ and appearance feature $\mathbf{P}^a(\mathbf{x}_i)$ of the P branch.

For the hybrid hash H branch, due to the finite-capacity nature of hash encoding, using three hash planes at high-resolution levels would duplicate the hash tables over three projections and increase the number of parameters compared with a single 3D hash grid at the same level. Therefore, we adopt coarse 2D hash planes and a fine 3D hash grid in the H branch. The coarse 2D hash planes reduce coarse-scale hash collisions, while the fine 3D hash grid preserves local detail representation and avoids the parameter growth caused by three high-resolution hash planes. For a sampled 3D point $\mathbf{x}_i=(x_i,y_i,z_i)$, the H-branch feature is defined as
\begin{equation}
\begin{gathered}
\mathbf{H}_c(\mathbf{x}_i)=
\operatorname{Concat}
\left(
\mathbf{E}_{yz}^{c}(y_i,z_i),
\mathbf{E}_{xz}^{c}(x_i,z_i),
\mathbf{E}_{xy}^{c}(x_i,y_i)
\right),\\
\mathbf{H}_f(\mathbf{x}_i)=
\mathbf{E}_{xyz}^{f}(x_i,y_i,z_i),\\
\mathbf{H}(\mathbf{x}_i)=
\operatorname{Concat}
\left(
\mathbf{H}_c(\mathbf{x}_i),
\mathbf{H}_f(\mathbf{x}_i)
\right).
\end{gathered}
\label{eq:h_branch_hash_feature}
\end{equation}
where $\mathbf{E}_{yz}^{c}$, $\mathbf{E}_{xz}^{c}$, and $\mathbf{E}_{xy}^{c}$ denote the coarse 2D hash planes queried by bilinear interpolation, while $\mathbf{E}_{xyz}^{f}$ denotes the fine 3D hash grid queried by trilinear interpolation. The coarse and fine hash features are concatenated to form the final H-branch feature $\mathbf{H}(\mathbf{x}_i)$.

\subsection{Multi-output Representation Architecture}
\label{subsec:multi_output_representation_architecture}

Existing NeRF-centric SLAM systems often rely on TSDF-based rendering for fast convergence, while density-based rendering provides another way to describe ray termination. To further integrate geometry and rendering constraints within the compact P-H representation without introducing additional networks, we design a compact multi-output representation architecture inspired by~\cite{hedman2021baking}. The unified decoder directly predicts TSDF, volume density, and color from the shared P-H hybrid features, enabling the TSDF-derived and density-derived ray termination distributions to be explicitly measured and aligned through the SKL loss in Sec.~\ref{subsec:objective_functions}.

Using the features defined in Sec.~\ref{subsec:compact_ph_hybrid_representation}, we construct the geometry input by concatenating the geometry feature $\mathbf{P}^g(\mathbf{x}_i)$, the hash feature $\mathbf{H}(\mathbf{x}_i)$, and the one-blob coordinate embedding $\gamma(\mathbf{x}_i)$. The geometry decoder $f_{\mathrm{tsdf}}$ predicts the TSDF value $s_i$ and a latent geometry code $\mathbf{g}_i$:
\begin{equation}
\begin{gathered}
\mathbf{z}_i^g=
\operatorname{Concat}
\left(
\mathbf{P}^g(\mathbf{x}_i),
\mathbf{H}(\mathbf{x}_i),
\gamma(\mathbf{x}_i)
\right),\\
f_{\mathrm{tsdf}}(\mathbf{z}_i^g)
\rightarrow
(s_i,\mathbf{g}_i).
\end{gathered}
\label{eq:tsdf_geometry_decoder}
\end{equation}
For density and color prediction, we concatenate the appearance feature $\mathbf{P}^a(\mathbf{x}_i)$, the one-blob embedding $\gamma(\mathbf{x}_i)$, and the latent geometry code $\mathbf{g}_i$. The density-color decoder $f_{\sigma c}$ predicts the volume density $\sigma_i$ and color $\bm{c}_i$:
\begin{equation}
\begin{gathered}
\mathbf{z}_i^a=
\operatorname{Concat}
\left(
\mathbf{P}^a(\mathbf{x}_i),
\gamma(\mathbf{x}_i),
\mathbf{g}_i
\right),\\
f_{\sigma c}(\mathbf{z}_i^a)
\rightarrow
(\sigma_i,\bm{c}_i).
\end{gathered}
\label{eq:density_color_decoder}
\end{equation}
In \equatoref{eq:tsdf_geometry_decoder} and \equatoref{eq:density_color_decoder}, $s_i$ is used for surface reconstruction, while $\sigma_i$ and $\bm{c}_i$ for volume rendering. The latent code $\mathbf{g}_i$ transfers geometry information from the TSDF branch to the density-color branch. In addition, the TSDF-density self-supervision is implemented by aligning their ray termination distributions, as detailed in Sec.~\ref{subsec:objective_functions}.

\subsection{Volume Rendering}
\label{subsec:volume_rendering}

Following NeRF-style differentiable rendering, we define a ray as $\mathbf{x}(z)=\mathbf{o}+z\mathbf{r}$, where $\mathbf{o}$ is the camera center, $\mathbf{r}$ is the ray direction, and $z$ denotes the sampled depth. We adopt a surface-aware sampling strategy, where $N$ points are stratified along each ray and additional samples are uniformly drawn around the observed depth. For all sampled points, the multi-output representation architecture predicts the TSDF value $s_i$ and color $\bm{c}_i$. Then, a TSDF-based neural rendering function converts $s_i$ into the opacity value used for volume accumulation:
\begin{equation}
\sigma(s_i)=(1/\beta)\operatorname{Sigmoid}(-s_i/\beta),
\label{eq:tsdf_to_density_mapping}
\end{equation}
where $\beta$ is a learnable parameter that controls the sharpness of the surface boundary. Based on the opacity values, the rendering weight of the $i$-th sampled point is computed as
\begin{equation}
\begin{gathered}
w_i=
T_i\left(1-\exp(-\sigma(s_i)\delta_i)\right),\quad
T_i=
\exp\left(
-\sum_{j=1}^{i-1}\sigma(s_j)\delta_j
\right).
\end{gathered}
\label{eq:tsdf_rendering_weight}
\end{equation}
where $\delta_i=z_{i+1}-z_i$ denotes the distance interval between adjacent samples. The rendered color and depth are obtained by weighted accumulation:
\begin{equation}
\begin{gathered}
\hat{C}=
\sum_{i=1}^{N}w_i c_i, \quad 
\hat{D}=
\sum_{i=1}^{N}w_i z_i.
\end{gathered}
\label{eq:rendered_color_depth}
\end{equation}

\subsection{Complementary Overlap Window Optimization}
\label{subsec:complementary_overlap_window_optimization}

\subsubsection{Sliding-window Selection Strategy}
\label{subsubsec:sliding_window_selection_strategy}
At the observation level, reuse of historical observations is essential for exploiting a compact scene representation. Existing NeRF-based SLAM systems~\cite{zhu2022niceslam,wang2023coslam,deng2024plgslam} rely on overlap-based local selection, random historical sampling, or global keyframe sampling. However, overlap-dominated windows may lack historical diversity, whereas weakly related historical frames may interfere with local geometric optimization. We therefore construct complementary temporal constraints by jointly considering recent pose continuity, local geometric overlap, and long-term historical supervision.

We design a complementary overlap-based sliding-window selection strategy to construct such an optimization window. Let $\mathcal{W}$ denote the target number of selected keyframes in the optimization window, and the current frame is optimized together with these selected keyframes. Thus, the effective optimization unit contains $\mathcal{W}$ keyframes and the current frame. Instead of relying on a single source of historical observations, our method decomposes the window into five complementary subsets:

\begin{equation}
\begin{gathered}
\mathcal{K}_{t}^{\mathrm{win}}=
\operatorname{Unique}
\left(
\mathcal{K}_{t}^{\mathrm{rec}}
\Vert
\mathcal{K}_{t}^{\mathrm{top}}
\Vert
\mathcal{K}_{t}^{\mathrm{lrand}}
\Vert
\mathcal{K}_{t}^{\mathrm{hfix}}
\Vert
\mathcal{K}_{t}^{\mathrm{hrand}}
\right),
\left|
\mathcal{K}_{t}^{\mathrm{win}}
\right|
\leq
\mathcal{W},
\end{gathered}
\label{eq:complementary_window_composition}
\end{equation}
where the five subsets denote recent keyframes, top-overlap local keyframes, randomized high-overlap local keyframes, fixed historical keyframes, and random historical keyframes, respectively. Here, $\Vert$ denotes ordered concatenation, and $\operatorname{Unique}(\cdot)$ removes duplicates while preserving the semantic order.

We first retain $N_r$ most recent keyframes as temporal anchors:
\begin{equation}
\left|
\mathcal{K}_{t}^{\mathrm{rec}}
\right|
=
N_r.
\label{eq:recent_keyframe_quota}
\end{equation}
The remaining budget is divided into local and historical parts:
\begin{equation}
\begin{gathered}
N_{\mathrm{rem}}=\mathcal{W}-N_r,\quad
N_l=\left\lfloor N_{\mathrm{rem}}/2 \right\rfloor,\quad
N_h=N_{\mathrm{rem}}-N_l,
\end{gathered}
\label{eq:local_history_quota}
\end{equation}
where $N_l$ and $N_h$ are the quotas for local and historical keyframes, respectively.

For local keyframe selection, candidates are first restricted to the current local radiance-field region to avoid unstable constraints from distant or cross-region keyframes. The most recent keyframes are excluded from the overlap candidate set because they have already been included in $\mathcal{K}_{t}^{\mathrm{rec}}$. We estimate the visual overlap by sampling 3D points along rays in the current frame and projecting them into each candidate keyframe:
\begin{equation}
O(t,k)=
\sum_{\mathbf{x}\in\mathcal{X}_{t}}
\mathbb{I}
\left(
\pi_k(\mathbf{x})\in\Omega_e
\right),
\label{eq:overlap_score}
\end{equation}
where $\mathcal{X}_{t}$ denotes the sampled 3D points from the current frame, $\pi_k(\cdot)$ is the projection function of candidate keyframe $k$, $\Omega_e$ is the valid image region after removing boundary margins, and $\mathbb{I}(\cdot)$ is the indicator function.

The local quota is further divided into a top-overlap subset and a randomized high-overlap subset:
\begin{equation}
\begin{gathered}
N_{\mathrm{top}}=\operatorname{round}(\rho N_l),\quad
N_{\mathrm{lrand}}=N_l-N_{\mathrm{top}},
\end{gathered}
\label{eq:local_top_random_quota}
\end{equation}
where $\rho$ is the top-overlap ratio. The subset $\mathcal{K}_{t}^{\mathrm{top}}$ contains the $N_{\mathrm{top}}$ keyframes with the highest overlap scores. The subset $\mathcal{K}_{t}^{\mathrm{lrand}}$ is randomly sampled from a high-overlap candidate pool rather than from all historical keyframes, which preserves strong visual overlap while improving local window diversity.

For historical keyframe selection, candidates are also restricted to the current local radiance-field region. After removing recent and local-overlap keyframes, the historical quota is divided into fixed and random subsets:
\begin{equation}
\begin{gathered}
\left|
\mathcal{K}_{t}^{\mathrm{hfix}}
\right|
=
N_{\mathrm{hfix}},\quad
\left|
\mathcal{K}_{t}^{\mathrm{hrand}}
\right|
=
N_h-N_{\mathrm{hfix}}.
\end{gathered}
\label{eq:history_keyframe_quota}
\end{equation}
The fixed historical keyframes are selected by temporally uniform sampling from the RF-constrained history pool, providing stable long-term constraints. The remaining historical keyframes are randomly sampled from the same filtered pool to increase historical diversity and reduce local representation forgetting.


\subsubsection{Keyframe Selection Strategy}
\label{subsubsec:keyframe_selection_strategy}

For sliding-window optimization, keyframe selection is also crucial for providing effective candidate observations. Existing SLAM systems commonly adopt fixed-interval, random, or overlap-based keyframe strategies~\cite{zhu2022niceslam,wang2023coslam,deng2024plgslam}, which may overlook frames where tracking quality degrades or new viewpoints provide useful constraints.
Therefore, we adopt a new additive keyframe selection strategy. Specifically, in addition to periodic keyframe insertion, the tracker further checks the tracking quality after each frame is processed.
If the total tracking loss is larger than the keyframe insertion threshold $\tau_{\mathrm{kf}}$, the current frame is inserted into the keyframe queue as an additional keyframe.
To avoid overly frequent insertion caused by short-term loss fluctuations, we impose a minimum trigger interval of $2$ frames.

In addition, we introduce a reprojection-overlap criterion to detect frames with insufficient visual overlap with existing keyframes. When the reprojection overlap of the current frame is lower than the threshold, the frame is treated as a useful new observation and added to the keyframe queue. This strategy improves mapping stability by preserving regular keyframe coverage while supplementing frames with poor tracking quality or low overlap.

\subsubsection{Loss-aware Bundle Adjustment}
\label{subsubsec:loss_aware_bundle_adjustment}

The complementary overlap window provides balanced keyframe constraints, while its effect also depends on when pose--map optimization is activated. Existing NeRF-centric SLAM frameworks often adopt periodic or global bundle adjustment to reduce camera drift~\cite{zhu2022niceslam,wang2023coslam,deng2024plgslam}, but fixed BA scheduling may not respond promptly to abrupt tracking degradation. Therefore, we introduce a loss-aware BA strategy as a supplement to the regular mapping and global BA schedule.

Specifically, while preserving periodic BA, a non-periodic BA is activated when the tracking loss first exceeds the BA trigger threshold $\tau_{\mathrm{BA}}$. The threshold $\tau_{\mathrm{BA}}$ is set slightly larger than the keyframe insertion threshold $\tau_{\mathrm{kf}}$, so that BA is triggered only under more severe tracking degradation. To avoid frequent additional optimization, we use a cooldown window of $10$ frames. Within this window, another non-periodic BA is triggered only when the loss of a new high-loss frame exceeds the previously triggered loss by a margin $\delta_{\mathrm{BA}}$. This design preserves the regular BA schedule while providing timely pose--map correction under unstable tracking.

\subsection{Two-stage Neural Tracking}
\label{subsec:two_stage_neural_tracking}

Camera tracking in NeRF-centric SLAM is commonly formulated as pose optimization through neural rendering losses. However, such optimization often has a limited convergence basin and lacks explicit feature correspondences, making it sensitive to initialization errors~\cite{cartillier2024slaim}. Some recent systems introduce classical visual SLAM components to improve tracking robustness~\cite{huang2024photoslam,peng2024rtgslam,chung2023orbeezslam}, but they usually rely on a relatively complete classical SLAM pipeline, which increases system coupling and implementation complexity.

Different from these systems, we only encapsulate ORB-based tracking and geometric pose estimation as a lightweight initialization module. Specifically, at frame $0$, the ORB module is initialized together with the initial scene coordinate system; at frame $1$, the ORB result is not used, and the pose of frame $0$ is directly copied as the initialization; from frame $2$ onward, the ORB-estimated pose is used as the initial pose for neural tracking. The neural tracker then refines the camera rotation and translation by minimizing the neural rendering loss.


\subsection{Objective Functions}
\label{subsec:objective_functions}

We apply four loss terms to jointly optimize the scene representation, the MLP decoders, and camera poses, including the color loss, depth loss, TSDF loss, and self-supervised consistency loss. Given a set of sampled rays $\mathcal{R}$, we use the L2 loss to measure the difference between the rendered color/depth and the corresponding RGB-D observations:
\begin{equation}
\mathcal{L}_{c}
=
\frac{1}{|\mathcal{R}|}
\sum_{r\in\mathcal{R}}
\left\|
\hat{\mathbf{C}}(r)-\mathbf{C}(r)
\right\|_{2}^{2},
\label{eq:color_loss}
\end{equation}
\begin{equation}
\mathcal{L}_{d}
=
\frac{1}{|\mathcal{R}|}
\sum_{r\in\mathcal{R}}
\left(
\hat{D}(r)-D(r)
\right)^{2},
\label{eq:depth_loss}
\end{equation}

Following~\cite{johari2023eslam}, we further impose sample-wise TSDF supervision along each ray. For a sampled point $p$ with sampled depth $d_p$, the free-space region is supervised to have a truncated TSDF value close to $1$:
\begin{equation}
\mathcal{L}_{fs}
=
\frac{1}{|\mathcal{R}|}
\sum_{r\in\mathcal{R}}
\frac{1}{|\mathcal{N}_{fs}(r)|}
\sum_{p\in\mathcal{N}_{fs}(r)}
\left(
s(p)-1
\right)^{2},
\label{eq:free_space_loss}
\end{equation}
where $\mathcal{N}_{fs}(r)$ denotes the sampled points outside the truncation band. For points near the observed surface, we split the truncation region into a middle region and a tail region:
\begin{equation}
\mathcal{L}_{mid}
=
\frac{1}{|\mathcal{R}|}
\sum_{r\in\mathcal{R}}
\frac{\lambda_{mid}}{|\mathcal{N}_{mid}(r)|}
\sum_{p\in\mathcal{N}_{mid}(r)}
\left(
s(p)+D(r)-d_p
\right)^{2},
\label{eq:mid_tsdf_loss}
\end{equation}
\begin{equation}
\mathcal{L}_{tail}
=
\frac{1}{|\mathcal{R}|}
\sum_{r\in\mathcal{R}}
\frac{\lambda_{tail}}{|\mathcal{N}_{tail}(r)|}
\sum_{p\in\mathcal{N}_{tail}(r)}
\left(
s(p)+D(r)-d_p
\right)^{2}.
\label{eq:tail_tsdf_loss}
\end{equation}
Here, $\mathcal{N}_{mid}(r)$ contains points within the central truncation band $[D(r)-0.4\delta,D(r)+0.4\delta]$, while $\mathcal{N}_{tail}(r)$ contains the remaining near-surface points within the truncation region. The TSDF loss is then defined as
\begin{equation}
\mathcal{L}_{sdf}
=
\mathcal{L}_{fs}
+
\mathcal{L}_{mid}
+
\mathcal{L}_{tail}.
\label{eq:tsdf_loss}
\end{equation}

To keep the TSDF-based reconstruction branch consistent with the density-based rendering branch, we introduce the self-supervised consistency loss in~\cite{feng2026mhedslam}. The TSDF-derived ray termination distribution is given by the rendering weight $w_i$ in \equatoref{eq:tsdf_rendering_weight}. Meanwhile, the density $\sigma_i$ predicted by the density-color decoder in \equatoref{eq:density_color_decoder} is converted into a density-derived ray termination distribution:
\begin{equation}
\begin{gathered}
\hat{w}_i
=
\hat{\alpha}_i
\prod_{j=1}^{i-1}
\left(
1-\hat{\alpha}_j
\right),\quad 
\hat{\alpha}_i
=
1-\exp(-\sigma_i\delta_i),
\end{gathered}
\label{eq:density_ray_termination_weight}
\end{equation}
We then use the symmetric KL divergence to encourage the TSDF-derived distribution $w$ and the density-derived distribution $\hat{w}$ to be consistent:
\begin{equation}
\mathcal{L}_{s}
=
\frac{1}{|\mathcal{R}|}
\sum_{r\in\mathcal{R}}
\frac{1}{2}
\sum_{i=1}^{N}
\left[
w_i
\log
\frac{w_i+\epsilon}{\hat{w}_i+\epsilon}
+
\hat{w}_i
\log
\frac{\hat{w}_i+\epsilon}{w_i+\epsilon}
\right],
\label{eq:self_supervised_consistency_loss}
\end{equation}
where $\epsilon$ is a small constant for numerical stability.

The final objective function is formulated as
\begin{equation}
\mathcal{L}
=
\lambda_c\mathcal{L}_{c}
+
\lambda_d\mathcal{L}_{d}
+
\lambda_{sdf}\mathcal{L}_{sdf}
+
\lambda_s\mathcal{L}_{s},
\label{eq:overall_loss}
\end{equation}

\section{Experiments}
\label{sec:experiments}

\subsection{Experimental Setup}
\label{subsec:experimental_setup}

\subsubsection{Datasets}
\label{subsubsec:datasets}

We evaluate CHOW-SLAM on three widely used RGB-D SLAM benchmarks, including both synthetic and real-world indoor scenes. Following the evaluation protocols of iMAP~\cite{sucar2021imap}, NICE-SLAM~\cite{zhu2022niceslam}, and Co-SLAM~\cite{wang2023coslam}, we assess reconstruction and tracking performance on eight synthetic scenes from Replica~\cite{straub2019replica}. For real-world camera tracking, we use six indoor sequences from ScanNet~\cite{dai2017scannet}. In addition, three standard sequences from TUM RGB-D~\cite{sturm2012benchmark}, namely fr1/desk, fr2/xyz, and fr3/office, are used to evaluate tracking robustness on real RGB-D data with accurate ground-truth trajectories.

\subsubsection{Metrics}
\label{subsubsec:metrics}

For reconstruction evaluation, we report Depth L1 (cm), Accuracy (cm), Completion (cm), and Completion Ratio (\%) with a distance threshold of $5$ cm. Following Co-SLAM~\cite{wang2023coslam}, we evaluate reconstructed meshes after removing unobserved regions outside all camera frustums and applying mesh culling to discard redundant geometry within the camera view but outside the valid scene region. For tracking, we use the ATE RMSE (cm)~\cite{sturm2012benchmark}. We also report the running speed in Frames Per Second (FPS), computed following the protocol of Co-SLAM~\cite{wang2023coslam}. Unless otherwise specified, bold values indicate the best result for each metric within the comparison group.

\subsubsection{Baselines}
\label{subsubsec:baselines}

We compare our system with representative RGB-D SLAM methods in terms of reconstruction quality and camera tracking accuracy. The baselines include NeRF-based methods: iMAP~\cite{sucar2021imap}, NICE-SLAM~\cite{zhu2022niceslam}, Co-SLAM~\cite{wang2023coslam}, and QQ-SLAM~\cite{jiang2025qqslam}; 3DGS-based methods: MonoGS~\cite{matsuki2024gaussianslam} and SplaTAM~\cite{keetha2024splatam}; and classical SLAM methods: Kintinuous~\cite{whelan2015kintinuous}, BAD-SLAM~\cite{schops2019badslam}, and ORB-SLAM2~\cite{murartal2017orbslam2}. Since 3DGS-based methods often produce less explicit surfaces and make mesh extraction less reliable, we only compare their camera tracking results for fairness.

\subsubsection{Implementation Details}
\label{subsubsec:implementation_details}

We run CHOW-SLAM on a desktop PC with an Intel Core i7-12700K CPU and an NVIDIA RTX 3090 GPU. Unless otherwise specified, all experiments use the same compact hybrid representation. The hash branch uses a 16-level HashGrid encoding, with the first 8 levels represented by three orthogonal 2D hash planes and the remaining 8 levels by a 3D hash grid. The auxiliary explicit branch consists of a coarse 3D grid and fine tri-planes. The geometry branch uses resolutions of $24$ cm and $6$ cm for the coarse grid and fine planes, while the appearance branch uses $24$ cm and $3$ cm, respectively. Each explicit feature level has 32 channels, resulting in a 64-channel auxiliary feature after concatenation. We use One-blob positional encoding with 16 bins. Both the TSDF/SDF decoder and the color-density decoder are two-layer MLPs with a hidden dimension of 32.

In the mapping thread, we perform 10 iterations with 2048 sampled rays per iteration on Replica~\cite{straub2019replica}, ScanNet~\cite{dai2017scannet}, and 20 iterations on TUM RGB-D~\cite{sturm2012benchmark}. In the tracking thread, all datasets use 10 iterations. Each tracking iteration samples 1024 rays on Replica, and TUM RGB-D, and 2048 rays on ScanNet. For surface-aware ray sampling, Replica use $N_u=32$ stratified samples and $N_s=8$ surface-guided samples per ray, while ScanNet and TUM RGB-D use $N_u=48$ and $N_s=8$.

The mapping loss weights are $\lambda_{fs}=5$, $\lambda_{center}=200$, $\lambda_{tail}=10$, $\lambda_d=0.1$, and $\lambda_c=5$, while the tracking loss weights are $\lambda_{fs}=10$, $\lambda_{center}=200$, $\lambda_{tail}=50$, $\lambda_d=1$, and $\lambda_c=5$. The SKL loss is enabled with an initial weight of $\lambda_S=5\times10^{-4}$ and linearly decays to 0 with the number of
iterations. 
For loss-aware keyframe insertion and BA scheduling, $\tau_{\mathrm{kf}}$ is set to 2.2, 1.2, and 1.3 for Replica, ScanNet, and TUM RGB-D, respectively, while $\tau_{\mathrm{BA}}$ is set slightly larger than $\tau_{\mathrm{kf}}$. The margin $\delta_{\mathrm{BA}}$ is set to 0.2, 0.15, and 0.6 for these three datasets, respectively.

For complementary overlap-window selection, we set the window size to $\mathcal{W}=20$, the number of recent keyframes to $N_r=2$, the top-overlap ratio to $\rho=0.4$, and the number of fixed historical keyframes to $N_{\mathrm{hfix}}=2$.

\subsection{Experimental results}
\subsubsection{Evaluation on Replica}
\label{subsubsec:evaluation_on_replica}

\begin{figure}[!htbp]
\centering
\includegraphics[width=0.8\linewidth]{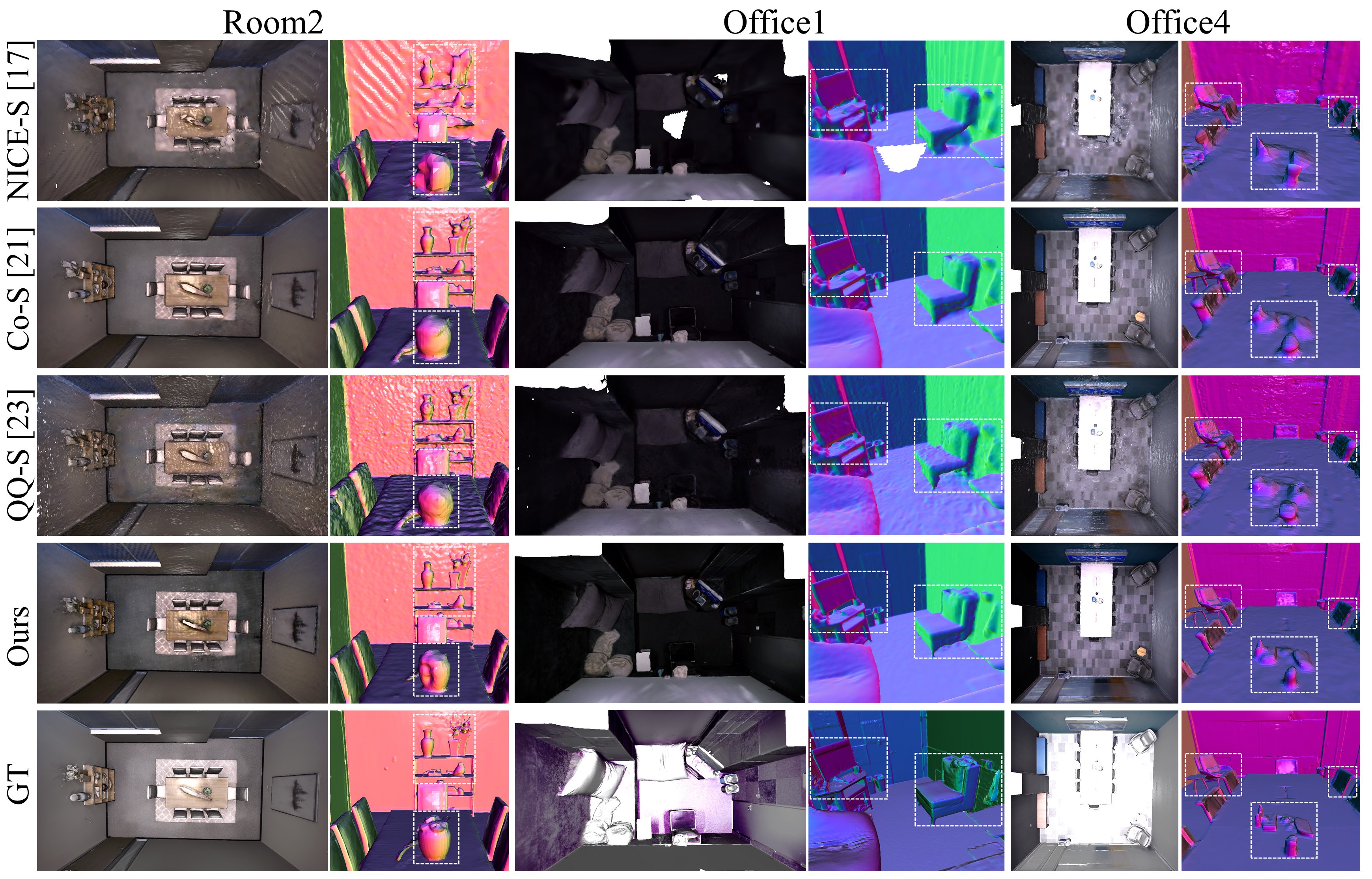}
\caption{Qualitative reconstruction comparison with state-of-the-art NeRF-based dense RGB-D VSLAM systems on Replica~\cite{straub2019replica}. Our system reconstructs more complete and smoother scene geometry, while the white dashed boxes highlight clearer details around small objects and object boundaries.}
\label{fig:replica_reconstruction}
\end{figure}

We evaluate the reconstruction and tracking performance of our system on eight scenes from Replica~\cite{straub2019replica}, with the results reported in Table~\ref{tab:replica_reconstruction_tracking} and Table~\ref{tab:replica_tracking_comparison}. Compared with recent NeRF-based systems, our method achieves the best average performance on all reconstruction metrics, with a particularly clear advantage in Completion Ratio. In terms of average tracking error, our method outperforms all baselines except the 3DGS-based SplaTAM~\cite{keetha2024splatam}.

\begin{figure}[!htbp]
\centering
\includegraphics[width=0.8\linewidth]{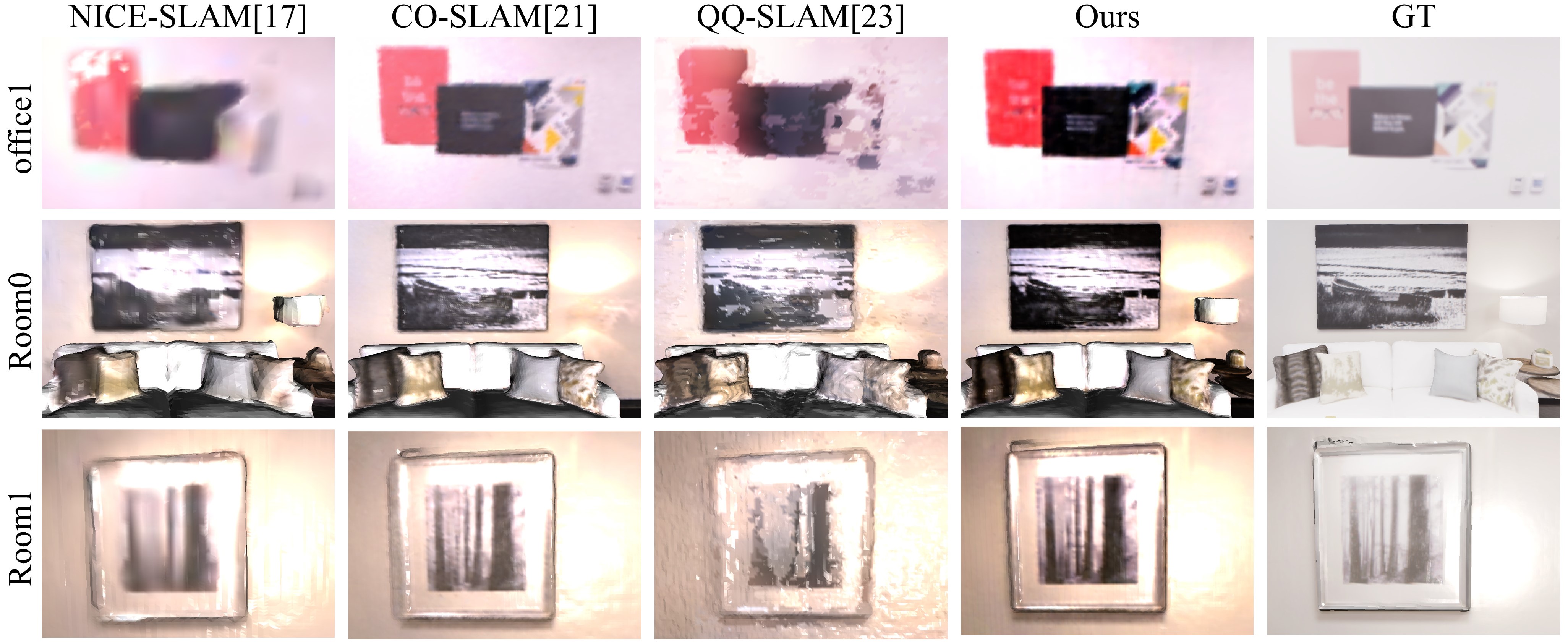}
\caption{Qualitative color-rendering comparison with state-of-the-art NeRF-based dense RGB-D VSLAM systems on Replica~\cite{straub2019replica}. Our system preserves sharper appearance details and produces more faithful colors closer to the ground truth.}
\label{fig:replica_rendering}
\end{figure}

The qualitative comparisons in Fig.~\ref{fig:replica_reconstruction} and Fig.~\ref{fig:replica_rendering} further support the quantitative results. As shown in Fig.~\ref{fig:replica_reconstruction}, other methods either produce incomplete and over-smoothed geometry with lost high-frequency details or introduce noisy artifacts around fine structures and surface transitions. In contrast, our method achieves more complete global geometry while preserving smooth surfaces and sharper local structures, resulting in reconstructions that are more faithful to the ground truth.
Fig.~\ref{fig:replica_rendering} further demonstrates the advantage of our method in color rendering, preserving sharper appearance details and producing more faithful rendered colors. These improvements benefit from the scale-aware complementary P-H representation, where the P branch preserves scene continuity and the H branch captures fine-scale geometry and appearance, reducing representation redundancy while suppressing hash-collision artifacts.

\begin{table}[!htbp]
\centering
\begin{threeparttable}
\caption{Reconstruction and tracking results on the Replica dataset.}
\label{tab:replica_reconstruction_tracking}
{\fontsize{8pt}{6pt}\selectfont
\setlength{\tabcolsep}{3pt}
\renewcommand{\arraystretch}{1.0}
\begin{tabular}{llccccc}
\toprule
\multirow{2}{*}{Scene}
& \multirow{2}{*}{Methods}
& \multicolumn{4}{c}{Reconstruction (cm)}
& \multicolumn{1}{c}{Localization (cm)} \\
\cmidrule(lr){3-6}
\cmidrule(lr){7-7}
& & Depth L1$\downarrow$ & Acc.$\downarrow$ & Comp.$\downarrow$
& Comp. ratio (\%)$\uparrow$ & ATE RMSE$\downarrow$ \\
\midrule

\multirow{5}{*}{Room0}
& iMAP~\cite{sucar2021imap} & 5.08 & 4.01 & 5.84 & 78.34 & 5.43 \\
& NICE-SLAM~\cite{zhu2022niceslam} & 1.79 & 2.44 & 2.60 & 91.81 & 1.69 \\
& Co-SLAM~\cite{wang2023coslam} & 1.05 & \textbf{2.11} & 2.02 & 95.26 & 0.72 \\
& QQ-SLAM~\cite{jiang2025qqslam} & 1.09 & 2.38 & \textbf{1.76} & 96.39 & 0.58 \\
& \textbf{Ours} & \textbf{0.72} & 2.33 & 1.80 & \textbf{96.72} & \textbf{0.48} \\
\midrule

\multirow{5}{*}{Room1}
& iMAP~\cite{sucar2021imap} & 3.44 & 3.04 & 4.40 & 85.85 & 3.22 \\
& NICE-SLAM~\cite{zhu2022niceslam} & 1.33 & 2.10 & 2.19 & 93.56 & 2.04 \\
& Co-SLAM~\cite{wang2023coslam} & 0.85 & \textbf{1.68} & 1.81 & 95.19 & 1.32 \\
& QQ-SLAM~\cite{jiang2025qqslam} & 0.69 & 2.62 & 1.77 & 95.49 & 1.16 \\
& \textbf{Ours} & \textbf{0.67} & 2.37 & \textbf{1.59} & \textbf{96.75} & \textbf{0.53} \\
\midrule

\multirow{5}{*}{Room2}
& iMAP~\cite{sucar2021imap} & 5.78 & 3.84 & 5.07 & 79.40 & 2.85 \\
& NICE-SLAM~\cite{zhu2022niceslam} & 2.20 & 2.17 & 2.73 & 91.48 & 1.55 \\
& Co-SLAM~\cite{wang2023coslam} & 2.37 & 1.99 & 1.96 & 93.58 & 1.27 \\
& QQ-SLAM~\cite{jiang2025qqslam} & 2.48 & 2.00 & 1.82 & 94.28 & 0.87 \\
& \textbf{Ours} & \textbf{1.03} & \textbf{1.59} & \textbf{1.59} & \textbf{96.93} & \textbf{0.36} \\
\midrule

\multirow{5}{*}{Office0}
& iMAP~\cite{sucar2021imap} & 3.79 & 3.34 & 3.62 & 83.59 & 2.60 \\
& NICE-SLAM~\cite{zhu2022niceslam} & 1.43 & 1.85 & 1.84 & 94.93 & 0.99 \\
& Co-SLAM~\cite{wang2023coslam} & 1.24 & 1.57 & 1.56 & 96.09 & 0.62 \\
& QQ-SLAM~\cite{jiang2025qqslam} & 1.18 & 1.55 & 1.57 & 96.10 & 0.52 \\
& \textbf{Ours} & \textbf{0.72} & \textbf{1.40} & \textbf{1.29} & \textbf{98.64} & \textbf{0.35} \\
\midrule

\multirow{5}{*}{Office1}
& iMAP~\cite{sucar2021imap} & 3.76 & 2.10 & 3.62 & 83.59 & 1.30 \\
& NICE-SLAM~\cite{zhu2022niceslam} & 1.58 & 1.56 & 1.82 & 94.11 & 0.90 \\
& Co-SLAM~\cite{wang2023coslam} & 1.48 & \textbf{1.31} & 1.59 & 94.65 & 0.52 \\
& QQ-SLAM~\cite{jiang2025qqslam} & \textbf{0.99} & 1.37 & 1.39 & 95.40 & 0.48 \\
& \textbf{Ours} & 1.14 & 1.50 & \textbf{1.19} & \textbf{97.92} & \textbf{0.40} \\
\midrule

\multirow{5}{*}{Office2}
& iMAP~\cite{sucar2021imap} & 3.97 & 4.06 & 4.73 & 79.73 & 5.94 \\
& NICE-SLAM~\cite{zhu2022niceslam} & 2.70 & 3.28 & 3.11 & 88.27 & 1.39 \\
& Co-SLAM~\cite{wang2023coslam} & 1.86 & 2.84 & 2.43 & 91.63 & 2.07 \\
& QQ-SLAM~\cite{jiang2025qqslam} & 1.76 & 3.43 & 2.14 & 94.07 & 1.74 \\
& \textbf{Ours} & \textbf{0.93} & \textbf{2.71} & \textbf{1.86} & \textbf{96.03} & \textbf{1.09} \\
\midrule

\multirow{5}{*}{Office3}
& iMAP~\cite{sucar2021imap} & 5.61 & 4.20 & 5.49 & 73.90 & 5.18 \\
& NICE-SLAM~\cite{zhu2022niceslam} & 2.10 & 3.01 & 3.16 & 87.68 & 3.97 \\
& Co-SLAM~\cite{wang2023coslam} & 1.66 & 3.06 & 2.72 & 90.72 & 1.47 \\
& QQ-SLAM~\cite{jiang2025qqslam} & 1.54 & 3.94 & 2.55 & 91.78 & 1.22 \\
& \textbf{Ours} & \textbf{0.81} & \textbf{2.57} & \textbf{1.97} & \textbf{96.01} & \textbf{0.59} \\
\midrule

\multirow{5}{*}{Office4}
& iMAP~\cite{sucar2021imap} & 5.71 & 4.34 & 6.65 & 74.77 & 2.42 \\
& NICE-SLAM~\cite{zhu2022niceslam} & 2.06 & 2.54 & 3.61 & 87.23 & 3.08 \\
& Co-SLAM~\cite{wang2023coslam} & 1.54 & 2.23 & 2.52 & 90.44 & 0.84 \\
& QQ-SLAM~\cite{jiang2025qqslam} & 1.68 & 2.16 & 2.46 & 91.53 & 0.73 \\
& \textbf{Ours} & \textbf{0.70} & \textbf{1.87} & \textbf{2.02} & \textbf{94.96} & \textbf{0.70} \\
\midrule

\multirow{5}{*}{Avg.}
& iMAP~\cite{sucar2021imap} & 4.64 & 3.62 & 4.93 & 80.50 & 3.62 \\
& NICE-SLAM~\cite{zhu2022niceslam} & 1.90 & 2.37 & 2.63 & 91.13 & 1.95 \\
& Co-SLAM~\cite{wang2023coslam} & 1.51 & 2.10 & 2.08 & 93.44 & 1.10 \\
& QQ-SLAM~\cite{jiang2025qqslam} & 1.42 & 2.43 & 1.93 & 94.38 & 0.91 \\
& \textbf{Ours} & \textbf{0.84} & \textbf{2.04} & \textbf{1.66} & \textbf{96.75} & \textbf{0.56} \\
\bottomrule
\end{tabular}
}

\begin{tablenotes}[flushleft]
\fontsize{8pt}{8pt}\selectfont
\setlength{\itemsep}{0pt}
\setlength{\labelsep}{0pt}
\item[]
We quantitatively compare our system with recent NeRF-based dense RGB-D VSLAM methods on the Replica~\cite{straub2019replica} scenes. Our method achieves the highest tracking accuracy and reaches state-of-the-art reconstruction quality.
\end{tablenotes}
\end{threeparttable}
\end{table}

\begin{table}[t]
\centering
\begin{threeparttable}
\caption{Tracking results of NeRF-based and 3DGS-based systems on the Replica dataset.}
\label{tab:replica_tracking_comparison}
{\fontsize{8pt}{8pt}\selectfont
\setlength{\tabcolsep}{3pt}
\renewcommand{\arraystretch}{1.0}
\begin{tabular}{llccccccccc}
\toprule
& Methods
& R0
& R1
& R2
& O0
& O1
& O2
& O3
& O4
& Avg. \\
\midrule

\multirow{5}{*}{NeRF}
& iMAP~\cite{sucar2021imap}
& 5.43
& 3.22
& 2.85
& 2.60
& 1.30
& 5.94
& 5.18
& 2.42
& 3.62 \\

& NICE-SLAM~\cite{zhu2022niceslam}
& 1.69
& 2.04
& 1.55
& 0.99
& 0.90
& 1.39
& 3.97
& 3.08
& 1.95 \\

& Co-SLAM~\cite{wang2023coslam}
& 0.72
& 1.32
& 1.27
& 0.62
& 0.52
& 2.07
& 1.47
& 0.84
& 1.10 \\

& QQ-SLAM~\cite{jiang2025qqslam}
& 0.58
& 1.16
& 0.87
& 0.52
& 0.48
& 1.74
& 1.22
& 0.73
& 0.91 \\

& Ours
& \textbf{0.48}
& \textbf{0.53}
& \textbf{0.36}
& \textbf{0.35}
& \textbf{0.40}
& \textbf{1.09}
& \textbf{0.59}
& \textbf{0.70}
& \textbf{0.56} \\
\midrule

\multirow{2}{*}{3DGS}
& MonoGS~\cite{matsuki2024gaussianslam}
& 0.76
& \textbf{0.37}
& \textbf{0.23}
& 0.66
& 0.72
& 0.30
& \textbf{0.19}
& 1.46
& 0.58 \\

& SplaTAM~\cite{keetha2024splatam}
& \textbf{0.31}
& 0.40
& 0.29
& \textbf{0.47}
& \textbf{0.27}
& \textbf{0.29}
& 0.32
& \textbf{0.55}
& \textbf{0.36} \\
\bottomrule
\end{tabular}
}
\begin{tablenotes}[flushleft]
\fontsize{8pt}{8pt}\selectfont
\setlength{\itemsep}{0pt}
\setlength{\labelsep}{0pt}
\item[]
For MonoGS, we report the single-process implementation, which performs additional mapping iterations. 
\end{tablenotes}

\end{threeparttable}
\end{table}

\begin{table}[!htbp]
\centering
\begin{threeparttable}
\caption{Tracking results of 3DGS-based and NeRF-based systems on ScanNet dataset.}
\label{tab:scannet_tracking}
{\fontsize{8pt}{8pt}\selectfont
\setlength{\tabcolsep}{5pt}
\renewcommand{\arraystretch}{1.0}
\begin{tabular}{llccccccc}
\toprule
& Methods & 0000 & 0059 & 0106 & 0169 & 0181 & 0207 & Avg. \\
\midrule
\multirow{6}{*}{NeRF}
& iMAP~\cite{sucar2021imap} & 55.95 & 32.06 & 17.50 & 70.51 & 32.10 & 11.91 & 36.67 \\
& NICE-SLAM~\cite{zhu2022niceslam} & 8.64 & 12.25 & 8.09 & 10.28 & 12.93 & 5.59 & 9.63 \\
& Co-SLAM~\cite{wang2023coslam} & 7.13 & 11.14 & 9.36 & 5.90 & 11.81 & 7.14 & 8.75 \\
& QQ-SLAM~\cite{jiang2025qqslam} & \textbf{6.99} & 9.47 & 8.82 & 6.48 & 13.30 & 5.86 & 8.49 \\
& Ours & 7.53 & \textbf{6.06} & \textbf{7.67} & \textbf{5.68} & \textbf{9.90} & \textbf{4.80} & \textbf{6.94} \\
\midrule
\multirow{2}{*}{3DGS}
& MonoGS~\cite{matsuki2024gaussianslam} & \textbf{9.58} & \textbf{6.16} & \textbf{7.05} & \textbf{10.66} & 18.23 & \textbf{7.46} & \textbf{9.86} \\
& SplaTAM~\cite{keetha2024splatam} & 12.81 & 10.13 & 17.68 & 12.08 & \textbf{11.14} & 7.49 & 11.89 \\
\bottomrule
\end{tabular}
}

\begin{tablenotes}[flushleft]
\fontsize{8pt}{8pt}\selectfont
\setlength{\itemsep}{0pt}
\setlength{\labelsep}{0pt}
\item[]
Our tracking results outperform those of the other systems on most scenes and achieve the lowest average error.
\end{tablenotes}

\end{threeparttable}
\end{table}

\subsubsection{Evaluation on ScanNet}
\label{subsubsec:evaluation_on_scannet}

Although Replica~\cite{straub2019replica} provides accurate reference geometry for quantitative reconstruction evaluation, its synthetic scenes cannot fully capture the structural complexity and sensing noise of real-world environments. We therefore further evaluate our system on ScanNet~\cite{dai2017scannet}, a large-scale real-world indoor dataset, to examine its robustness under practical conditions.

\begin{figure*}[!htbp]
\centering
\includegraphics[width=0.8\linewidth]{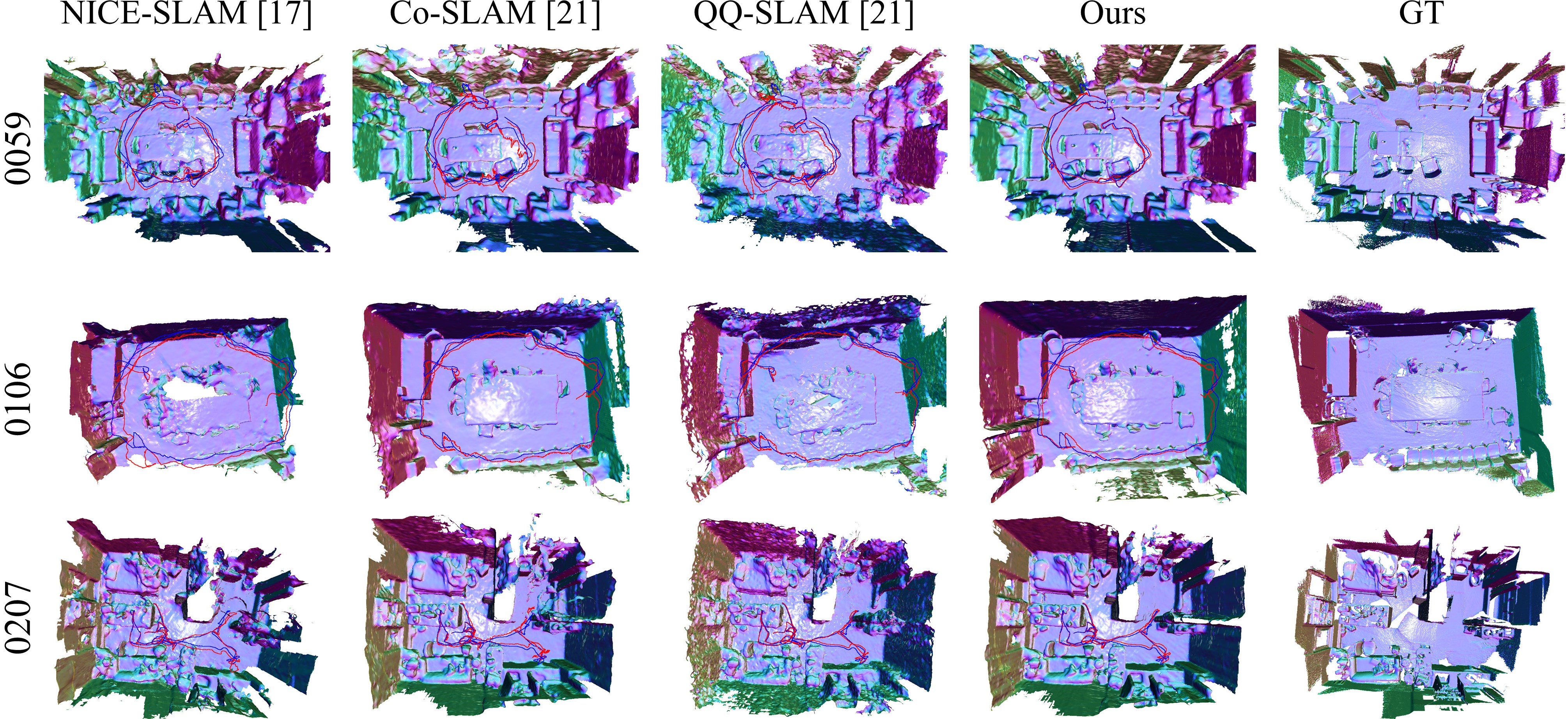}
\caption{Qualitative comparison with state-of-the-art NeRF-based dense RGB-D VSLAM systems on ScanNet~\cite{dai2017scannet}, focusing on reconstructed surface geometry and tracking accuracy. Blue and red curves denote the ground-truth and estimated camera trajectories, respectively. From the same viewpoints, our system reconstructs scene structures more consistent with the ground truth and estimates more accurate trajectories.}
\label{fig:scannet_qualitative}
\end{figure*}

Table~\ref{tab:scannet_tracking} presents a comparison between our system and representative NeRF-based and 3DGS-based methods. Our system achieves higher tracking accuracy on most sequences and obtains the lowest average ATE RMSE, demonstrating its robustness in real-world indoor environments. Fig.~\ref{fig:scannet_qualitative} compares the surface-normal renderings of the reconstructed meshes and camera trajectories on selected ScanNet scenes. The trajectories estimated by our system closely match the ground truth, while its reconstructions are more consistent with the ground truth, particularly in structural shape and local details. For example, in the upper-right restroom of scene0207, our system preserves a more complete and geometrically consistent structure, whereas the other methods produce noticeable deformation and missing surfaces. This improvement benefits from the complementary overlap window optimization, which provides effective and persistent temporal consistency constraints for online pose--map optimization.

\subsubsection{Evaluation on TUM RGB-D}
\label{subsubsec:evaluation_on_tum_rgbd}

Following~\cite{zhu2022niceslam,wang2023coslam,jiang2025qqslam}, we evaluate the tracking accuracy of our system on the TUM RGB-D dataset. As shown in Table~\ref{tab:tum_rgbd_tracking}, our system achieves strong performance on most sequences and obtains the best average accuracy among NeRF-based SLAM methods, with only a slightly higher error than MonoGS among NeRF-based and 3DGS-based methods. This shows that accurate dense mapping helps NeRF-based SLAM reduce the tracking gap to classical SLAM while preserving dense reconstruction capability.

\begin{table}[!htbp]
\centering
\begin{threeparttable}

\caption{Tracking results on the TUM RGB-D dataset.}
\label{tab:tum_rgbd_tracking}
{\fontsize{8pt}{8pt}\selectfont
\setlength{\tabcolsep}{8pt}
\renewcommand{\arraystretch}{1.0}
\begin{tabular}{llcccc}
\toprule
& Methods & fr1/desk & fr2/xyz & fr3/office & Avg. \\
\midrule
\multirow{3}{*}{Classic}
& Kintinuous~\cite{whelan2015kintinuous}   & 3.70 & 2.90 & 3.00 & 3.20 \\
& BAD-SLAM~\cite{schops2019badslam}     & 1.70 & 1.10 & 1.70 & 1.50 \\
& ORB-SLAM2~\cite{murartal2017orbslam2} & \textbf{1.60} & \textbf{0.40} & \textbf{1.00} & \textbf{1.00} \\
\midrule
\multirow{5}{*}{NeRF}
& iMAP~\cite{sucar2021imap}             & 4.90 & 2.00 & 5.80 & 4.23 \\
& NICE-SLAM~\cite{zhu2022niceslam}      & 2.85 & 2.16 & 3.14 & 2.72 \\
& Co-SLAM~\cite{wang2023coslam}         & 2.73 & 2.02 & 2.63 & 2.46 \\
& QQ-SLAM~\cite{jiang2025qqslam}        & 2.61 & 1.70 & 2.70 & 2.34 \\
& Ours                                  & \textbf{1.95} & \textbf{1.31} & \textbf{2.06} & \textbf{1.77} \\
\midrule
\multirow{2}{*}{3DGS}
& MonoGS~\cite{matsuki2024gaussianslam} & \textbf{1.55} & 1.58 & \textbf{1.65} & \textbf{1.59} \\
& SplaTAM~\cite{keetha2024splatam}      & 3.35 & \textbf{1.24} & 5.16 & 3.25 \\
\bottomrule
\end{tabular}}
\end{threeparttable}

\end{table}

\subsection{Ablation Study}
\label{subsec:ablation_study}

\begin{figure*}[!hbpt]
\centering
\includegraphics[width=0.8\linewidth]{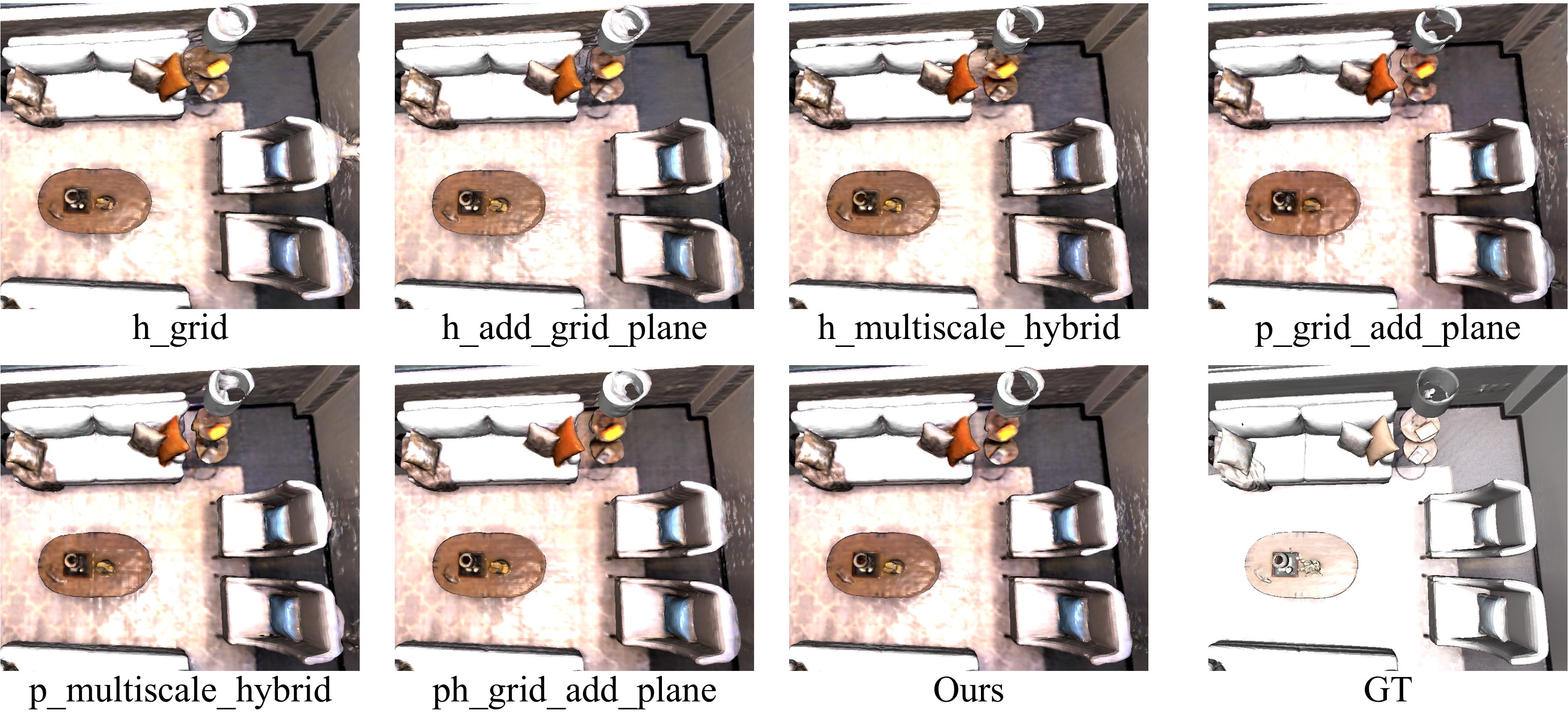}
\caption{Qualitative ablation of different scene encoding configurations on the Replica~\cite{straub2019replica} room1 scene.}
\label{fig:representation_ablation}
\end{figure*}

\begin{figure*}[!htbp]
\centering
\includegraphics[width=0.5\linewidth]{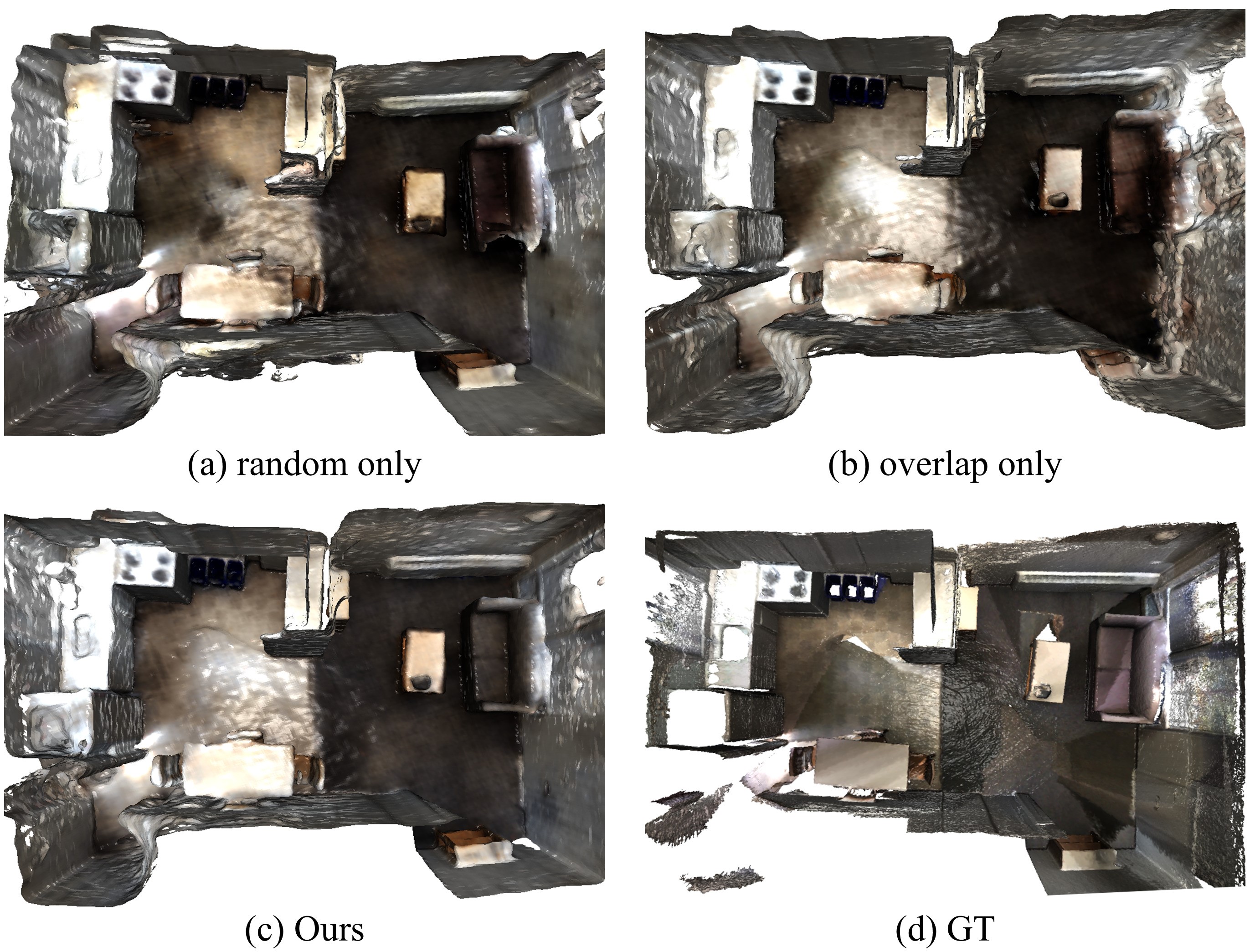}
\caption{Qualitative comparison of sliding-window selection strategies on ScanNet~\cite{dai2017scannet} scene0207.}
\label{fig:window_ablation}
\end{figure*}

We conduct ablation studies on the Compact P-H Hybrid Representation and Complementary Overlap Window Optimization (COWO), with the quantitative results reported in Table~\ref{tab:ablation_study}. Here, h and p denote the hybrid hash branch and the explicit parametric branch introduced in Sec.~\ref{subsec:compact_ph_hybrid_representation}, respectively. The suffix add indicates that grid and plane features are directly combined across all scales, whereas multiscale\_hybrid denotes the scale-aware hybrid configuration described in Sec.~\ref{subsec:compact_ph_hybrid_representation}.

For scene representation, the multi-scale hybrid design in the H branch yields clear overall improvements over the other hash configurations. In the P branch, although the multi-scale hybrid configuration performs slightly below direct grid--plane addition, it preserves most of the representation capability with substantially fewer parameters. The ph\_grid\_add\_plane configuration also achieves competitive performance, but directly combining grids and planes across all scales leads to excessive parameter growth. In contrast, our scale-aware P-H representation achieves better overall reconstruction and tracking performance while maintaining a compact parameter budget. For COWO, removing the sliding-window strategy, adaptive keyframe insertion, or loss-aware BA causes relatively small changes in reconstruction quality but a clear degradation in trajectory accuracy. Removing the complete COWO produces the largest tracking error, indicating that its components contribute more directly to accurate pose estimation by maintaining effective and persistent temporal constraints.

Fig.~\ref{fig:representation_ablation} visualizes the reconstructed meshes obtained with different encoding configurations on the Replica~\cite{straub2019replica} room1 scene. Although h\_multiscale\_hybrid avoids the artifact behind the chair, it preserves fewer details around the tabletop and lampshade in the upper-right region. In contrast, our method suppresses reconstruction artifacts while recovering these local structures more faithfully. This improvement results from combining the structured spatial support of the P branch with the fine-scale detail modeling capability of the H branch.

We further visualize different sliding-window selection strategies on ScanNet~\cite{dai2017scannet} scene0207 in Fig.~\ref{fig:window_ablation}. Random-only selection lacks sufficiently strong local geometric constraints and produces distorted structures, whereas overlap-only selection concentrates excessively on local observations and provides insufficient supervision for previously observed regions. By jointly incorporating high-overlap local observations and diverse historical keyframes, COWO produces a more complete and geometrically consistent reconstruction.

\begin{table}[t]
\centering
\begin{threeparttable}

\caption{Ablation study of the Compact P-H Hybrid Representation and Complementary Overlap Window Optimization.}
\label{tab:ablation_study}
{\fontsize{8pt}{8pt}\selectfont
\setlength{\tabcolsep}{4pt}
\renewcommand{\arraystretch}{1.0}
\begin{tabular}{clccccc}
\toprule
& \multirow{2}{*}{Name}
& \multicolumn{4}{c}{Reconstruction (cm)} & Localization (cm) \\
\cmidrule(lr){3-6}
\cmidrule(lr){7-7}
& 
& Depth L1 $\downarrow$
& Acc. $\downarrow$
& Comp. $\downarrow$
& Comp. Ratio (\%) $\uparrow$
& ATE RMSE $\downarrow$ \\
\midrule

(A) & h\_grid & 0.98 & 1.94 & 1.93 & 94.91 & 0.71 \\
(B) & h\_plane & 1.07 & 2.17 & 1.83 & 95.67 & 0.66 \\
(C) & h\_add\_grid\_plane & 1.03 & 1.98 & 1.77 & 95.71 & 0.61 \\
(D) & h\_multiscale\_hybrid & 0.92 & 2.24 & 1.79 & 96.12 & 0.60 \\
\midrule

(E) & p\_grid & 1.30 & 2.33 & 1.83 & 95.27 & 0.63 \\
(F) & p\_plane & 1.09 & 2.30 & 1.80 & 95.97 & 0.67 \\
(G) & p\_grid\_add\_plane & 0.99 & 2.25 & 1.68 & 96.52 & 0.59 \\
(H) & p\_multiscale\_hybrid & 1.01 & 2.30 & 1.71 & 96.33 & 0.62 \\
\midrule

(I) & ph\_grid\_add\_plane & 0.87 & \textbf{1.90} & 1.67 & 96.61 & 0.59 \\
\midrule

(J) & w/o sliding window & 0.88 & 2.09 & 1.68 & 96.57 & 0.72 \\
(K) & w/o adaptive keyframes & 0.88 & 2.08 & 1.69 & 96.68 & 0.64 \\
(L) & w/o loss-aware BA & 0.85 & 2.08 & 1.68 & 96.48 & 0.67 \\
(M) & w/o COWO & 0.90 & 2.21 & 1.73 & 95.97 & 0.75 \\
\midrule

& \textbf{Ours}
& \textbf{0.84}
& 2.04
& \textbf{1.66}
& \textbf{96.75}
& \textbf{0.56} \\
\bottomrule
\end{tabular}
}

\end{threeparttable}

\end{table}

\subsection{Performance Analysis}
\label{subsec:performance_analysis}
We evaluate the runtime efficiency and model parameters of NeRF-based and 3DGS-based SLAM systems on Replica~\cite{straub2019replica} and ScanNet~\cite{dai2017scannet}, as reported in Table~\ref{tab:runtime_model_complexity}. Our method achieves the second-highest runtime speed on both datasets, slightly below Co-SLAM~\cite{wang2023coslam} but faster than all other compared methods. Meanwhile, our method achieves more faithful scene reconstruction and more accurate camera tracking with a relatively small model size.

\begin{table}[!htbp]
\centering
\begin{threeparttable}
\caption{Time and memory analysis.}
\label{tab:runtime_model_complexity}
{\fontsize{8pt}{8pt}\selectfont
\setlength{\tabcolsep}{8pt}
\renewcommand{\arraystretch}{1.05}
\begin{tabular}{llcc}
\toprule
Dataset & Method & FPS $\uparrow$ & Model Param. (M) $\downarrow$ \\
\midrule

\multirow{6}{*}{Replica}
& NICE-SLAM~\cite{zhu2022niceslam} & 0.99 & 49.85 \\
& Co-SLAM~\cite{wang2023coslam} & \textbf{17.24} & \textbf{7.93} \\
& QQ-SLAM~\cite{jiang2025qqslam} & 7.08 & 7.96 \\
& MonoGS~\cite{matsuki2024gaussianslam} & 0.70 & 24.20 \\
& SplaTAM~\cite{keetha2024splatam} & 0.20 & 243.40 \\
& \textbf{Ours} & 12.53 & 14.72 \\
\midrule

\multirow{6}{*}{ScanNet}
& NICE-SLAM~\cite{zhu2022niceslam} & 0.73 & 85.85 \\
& Co-SLAM~\cite{wang2023coslam} & \textbf{6.41} & 9.97 \\
& QQ-SLAM~\cite{jiang2025qqslam} & 4.25 & 10.14 \\
& MonoGS~\cite{matsuki2024gaussianslam} & 2.30 & \textbf{5.60} \\
& SplaTAM~\cite{keetha2024splatam} & 0.20 & 156.40 \\
& \textbf{Ours} & 4.92 & 19.17 \\
\bottomrule
\end{tabular}}

\end{threeparttable}
\end{table}

\section{Conclusion}
\label{sec:conclusion}

This paper presents CHOW-SLAM, a compact and robust NeRF-based dense RGB-D SLAM system. First, our results demonstrate that the scale-aware combination of parametric and hash features can preserve fine geometry and appearance while avoiding redundant encoding parameters and hash-collision artifacts. Second, the Complementary Overlap Window Optimization provides effective and persistent temporal constraints by jointly exploiting recent, high-overlap local, and historical keyframes, leading to more accurate tracking and consistent mapping. In addition, classical ORB-based feature tracking and geometric pose estimation provide reliable pose initialization, further improving the stability of neural tracking.

The proposed representation and optimization principles provide a practical reference for researchers seeking to balance scene representation capability, model size, and persistent historical supervision in online neural SLAM.

\textbf{Limitations.} Our method remains limited in dynamic environments, outdoor and large-scale scenes, and texture-deficient regions. In future work, we will investigate dynamic-aware scene modeling and motion filtering, integrate loop-closure detection and optimization, and improve robustness in texture-deficient regions, with the goal of extending the system to more complex outdoor and long-term robotic applications.

\end{document}